# From Conception to Deployment: Intelligent Stroke Prediction Framework using Machine Learning and Performance Evaluation

Leila Ismail[1,2,*], Member, IEEE and Huned Materwala[1,2]

[1]*Intelligent Distributed Computing and Systems (INDUCE) Research Laboratory*
*Department of Computer Science and Software Engineering, College of Information Technology, United Arab Emirates University*
Al Ain, Abu Dhabi, 15551, United Arab Emirates

[2]*National Water and Energy Center*
*United Arab Emirates University*
Al Ain, Abu Dhabi, United Arab Emirates

*Correspondence: Leila Ismail (email: leila@uaeu.ac.ae)

*Abstract*—Stroke is the second leading cause of death worldwide. Machine learning classification algorithms have been widely adopted for stroke prediction. However, these algorithms were evaluated using different datasets and evaluation metrics. Moreover, there is no comprehensive framework for stroke data analytics. This paper proposes an intelligent stroke prediction framework based on a critical examination of machine learning prediction algorithms in the literature. The five most used machine learning algorithms for stroke prediction are evaluated using a unified setup for objective comparison. Comparative analysis and numerical results reveal that the Random Forest algorithm is best suited for stroke prediction.

## I. INTRODUCTION

Stroke is a neurological condition that causes the death of brain cells due to interrupted blood flow [1]. It is the second leading cause of death globally and accounts for 11% of total deaths[1]. According to the Centers for Disease Control and Prevention (CDC), more than 795,000 people in the United States suffer from stroke every year[2]. Between 2017 and 2018, approximately $53 billion in health expenditure was spent on stroke[2]. Prevalence of stroke is associated with the interaction of demographic, genetic, lifestyle, and medical conditions risk factors [1]. If not diagnosed and treated at an early stage, stroke can lead to visual, emotional, clinical, and communicable disorders [2]. To date, there are no simple means of treating and preventing clinical causes of stroke [1]. Machine learning classification algorithms have shown promising potential for the prediction of stroke.

Several works in the literature have evaluated the performance of different machine learning algorithms for stroke prediction using disparate experimental setups [3]–[8], making an objective comparison among these algorithms difficult. Furthermore, to the best of our knowledge, there exists no comprehensive framework that depicts the process of stroke prediction. In this paper, we propose an intelligent stroke prediction framework using machine learning that would support allied health professionals (doctors, pathologists, therapists, and medical technologists) for better diagnosis and prognosis [9] and will aid medical stakeholders (government, pharmaceutical companies, and health insurance providers) to develop an effective stroke prevention plan. The proposed framework is evaluated using the five most used machine learning approaches for stroke prediction in the literature [3]–[8] (Figure 1), Decision Tree (DT), Random Forest (RF), Support Vector Machine (SVM), Naïve Bayes (NB), and Logistic Regression (LR), to provide an objective comparison in a unified setup. The main contributions of this paper are as follows.

- We propose an intelligent stroke prediction framework that depicts the process of stroke data analytics from domain conception to model deployment.

- The mostly used machine learning approaches for stroke prediction are evaluated for objective comparison.

- We evaluate the impact of data balancing on the performance of machine learning approaches.

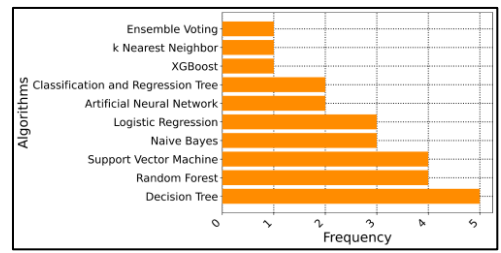

Fig. 1. Frequency of machine learning classification algorithms used in the literature for stroke prediction.

The rest of the paper is organized as follows: In section II, we present a summary of related work. Section III explains our proposed intelligent stroke prediction framework. The machine learning algorithms for stroke prediction are described in Section IV. The experimental setup, experiments, and numerical analysis are discussed in Section V. Section VI concludes the paper.

## II. RELATED WORK

Table I compares related work on machine learning-based stroke prediction. It shows that these works use different datasets and performance metrics for evaluation. In this paper, we propose an intelligent stroke prediction framework and evaluate it using the five most used machine learning algorithms in the literature, providing an objective comparison among these algorithms.

---

[1] https://www.who.int/news-room/fact-sheets/detail/the-top-10-causes-of-death (last accessed on April 25, 2022)

[2] https://www.cdc.gov/stroke/facts.htm (last accessed 25, 2022)

TABLE I.   WORK ON MACHINE LEARNING-BASED STROKE PREDICTION IN THE LITERATURE.

| Work | Dataset | Features | Observations | Algorithms evaluated | Evaluation metrics |
|---|---|---|---|---|---|
| [3] | Private – Mahasarakham hospital | Gender, marital status, smoking, alcohol, exercise, occupation, blood pressure, and cholesterol | 500 (*stroke*) and 500 (*no stroke*) | DT, SVM, NB, and KNN | Accuracy, precision*, recall*, and F-measure* |
| [4] | Private – emergency medical center of Chungnam National University Hospital, Korea | Level of consciousness, best gaze, visual, facial palsy, left and right motor arm, left and right motor leg, limb ataxia, sensory, best language, dysarthria, extinction and inattention, and left and right distal motor function | 227 | ANN, RF, XGBoost, SVM, NB, DT, LR, CART | Accuracy, precision, and recall |
| [5] | MarketScan Medicaid Multi-State Database | 4146 medications and medical conditions, age, and gender | 17,86 (*stroke*) and 10,800 (*no stroke*) | RF, SVM, DT, LR, and CART | AUC and training time |
| [6] | Stroke prediction dataset[3] | ID, gender, age, hypertension, heart disease, ever married, work type, residence type, average glucose, BMI, and smoking status | 249 (*stroke*) and 4,861 (*no stroke*) | RF, DT, LR, and voting | Accuracy, precision*, recall*, and F-measure* |
| [7] | Private | Brain MRI images | 192 | RF and SVM | Confusion matrix, accuracy, sensitivity, specificity, JI, and DSI |
| [8] | Private – faculty of Physical Therapy, Mahidol University, Thailand | Gender, age, province, marital status, education, and occupation | 250 (*stroke*) and 250 (*no stroke*) | ANN, NB, and DT | Accuracy, FPR, FNR, and AUC |
| This paper | Stroke prediction dataset[4] | ID, gender, age, hypertension, heart disease, ever married, work type, residence type, average glucose, BMI, and smoking status | 783 (*stroke*) and 42,617 (*no stroke*) | DT, RF, SVM, NB, and LR | Accuracy, precision*, recall*, and F-measure*, ROC, AUC, and execution times for hyperparameter tuning and model development and validation |

DT: Decision Tree; SVM: Support Vector Machine; NB: Naïve Bayes, KNN: K Nearest Neighbor; ANN: Artificial Neural Network; RF: Random Forest; LR: Logistic Regression; CART: Classification and Regression Tree; ROC: Receiver operating characteristic; AUC: Area Under the ROC Curve; BMI: Body Mass Index; FPR: False Positive Rate; FNR: False Negative Rate; *for both stroke and no stroke classes; JI: Jaccard Index; DSI: Dice Similarity Index

## III. PROPOSED INTELLIGENT STROKE PREDICTION FRAMEWORK

Figure 2 presents the stages involved in our proposed intelligent stroke prediction framework that is based on the data analytics lifecycle [10]. In the following subsections, we explain each stage in detail.

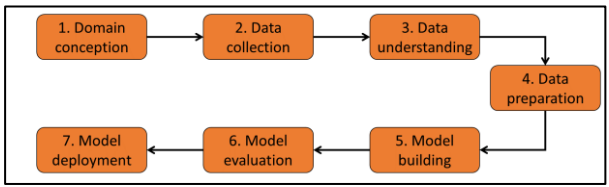

Fig. 2. Stages of the proposed intelligent stroke prediction framework.

### A. Domain Conception

In this stage, the stroke prediction problem is studied, i.e., ischemic or hemorrhagic stroke [1]. The potential stroke risk factors should be identified by surveying the literature [1] and/or consulting an expert. In addition, the objective of the prediction model should be stated clearly, i.e., binary class, post-stroke condition, or risk factor-specific predictions.

### B. Data Collection

In this stage, the stroke dataset is collected through public repositories such as the University of California Irvine (UCI) machine learning repository[5] and/or created using patients' medical records post-consent. The inclusion of stroke risk factors as features in the dataset should be ensured to develop an accurate and representative prediction model.

### C. Data Understanding

In this stage, the dataset should be aggregated, if it is scattered across multiple files. Furthermore, international disease coding systems should be referred to in case the risk factors are represented using a coding scheme such as the International Classification of Diseases (ICD)-9[6].

### D. Data Preparation

In this stage, the dataset should be preprocessed to remove irrelevant features, such as patient ID, to prevent the model from overfitting during its building stage. Moreover, outliers should be removed using machine learning approaches [11] and/or manual visualization to improve the prediction accuracy [12]. The categorical features (for instance, marital status) should be converted to numerical using techniques such as one-hot or label encoding [13]. Furthermore, the numerical risk factors should be normalized to avoid bias towards risk factors with a high range of values compared to the ones with a lower range [12]. The missing values in the dataset should be identified and treated by either removing the observations with missing values or imputing synthetic values using statistical or machine learning approaches [12]. The machine learning models can be developed using all the risk factors in the preprocessed dataset or a subset of factors selected using feature selection algorithms [14] and/or expert advice. In addition, the dataset should be balanced, if imbalanced, using over and under-sampling approaches [12].

### E. Model Building

In this stage, the preprocessed dataset is used to build the machine learning model. This can be either using the k-fold cross-validation [14] or the percentage split approach. Optimal hyperparameters' values for models should be used for building a model to improve prediction performance.

---

[3] https://www.kaggle.com/datasets/fedesoriano/stroke-prediction-dataset (last accessed on April 25, 2022)
[4] https://www.kaggle.com/datasets/shashwatwork/cerebral-stroke-predictionimbalaced-dataset (last accessed on April 25, 2022)
[5] https://archive.ics.uci.edu/ml/datasets.php (last accessed on April 25, 2022)
[6] https://www.cdc.gov/nchs/icd/icd9cm.htm (last accessed on April 25, 2022)

*F. Model Evaluation*

In this stage, the model is evaluated using the validation dataset. In the case of k-fold cross-validation, the model is validated using 1/k of the dataset during each fold. The model should be evaluated in terms of accuracy. In case the dataset is imbalanced, class-specific and macro average precision, recall, and F-measure, Receiver Operating Characteristic (ROC) curve, and Area under the ROC curve (AUC) should be evaluated. The complexity of the model should be evaluated by measuring the execution times for hyperparameter tuning, model building, and validation.

*G. Model Deployment*

In this stage developed validated model is used for stroke prediction by allied healthcare professionals. The model should be re-built based on updated and/or new observations in the dataset.

## IV. Machine Learning Algorithms for Stroke Prediction

In this section, the considered machine learning classification algorithms for stroke prediction are explained.

*A. Decision Tree*

DT develops a rule-based model by constructing a tree structure where each stroke risk factor represents a node [14]. The topmost node is known as the root node and the nodes with class labels (stroke and no stroke) are termed the leaf nodes. At each node, the algorithm traverses down to the next node/leaf by selecting the most informative risk factor using entropy-based Information gain or the Gini index. Given a stroke dataset with risk factors $\{R_1, R_2, ...\}$ and a stroke class label, information gain for a risk factor $R_1$ can be calculated as stated in Equation (1).

$$InfoGain_{R_1} = H_{stroke} - H_{stroke|R_1} \quad (1)$$

where $H_{stroke}$ and $H_{stroke|R_1}$ represent base (Equation 2) and conditional entropies of $R_1$ (Equation 3) respectively.

$$H_{stroke} = - \sum_{\forall s \in stroke} P(s) \log_2 P(s) \quad (2)$$

$$H_{stroke|R_1} = - \sum_{\forall r \in R_1} P(r) \sum_{\forall s \in stroke} P(s|r) \log_2 P(s|r) \quad (3)$$

where $s \in \{yes, no\}$ and $r$ is set of values for a risk factor $R_1$.

The Gini index (GI) can be calculated using Equation (4).

$$GI(R_1) = Gini(stroke) - \sum_{\forall r \in R} P(r).Gini(R_1 = r) \quad (4)$$

where $Gini(stroke)$ represents the Gini impurity of the stroke class as stated in Equation (5)

$$Gini(stroke) = 1 - \big(P(stroke = yes)\big)^2 - \big(P(stroke = no)\big)^2 \quad (5)$$

DT algorithm mainly consists of two hyperparameters: 1) splitting criteria and 2) the maximum number of features to be considered while making the split decision.

*B. Random Forest*

RF consists of multiple DTs, where each tree is created using a random sample of dataset and risk factors features [14]. It mainly involves three hyperparameters; 1) number of DTs, 2) splitting criteria for each DT, and 3) maximum features to be considered for a split.

*C. Support Vector Machine*

SVM constructs a hyperplane that separates n-dimensional risk factor feature space to partition the observations into stroke and no-stroke classes [14]. A hyperplane resulting in the maximum distance between observations of the two classes (referred to as margin) is selected. The cost function to maximize the margin is stated in Equation (6).

$$min_w \lambda ||w||^2 + \sum_{i=1}^{n}(1 - s_i <R_i, w>)_+ \quad (6)$$

where $\lambda$ is the regularization parameter, $w$ is normal to the hyperplane, and $n$ is the number of observations in the dataset.

SVM uses a kernel function (linear, polynomial, radial basis function, or sigmoid) to develop a non-linear hyperplane [15]. It consists of three hyperparameters; 1) kernel function, 2) regularization parameter, and 3) kernel coefficient.

*D. Naïve Bayes*

NB formulates the relationship between the probabilities and conditional probabilities of two risk factors $R_1$ and $R_2$ [14]. Given the prevalence of $R_1$ for an observation, the conditional probability of $R_2's$ occurrence is given by Equation (7).

$$P(R_2|R_1) = \frac{P(R_1 \cap R_2)}{P(R_1)} = \frac{P(R_1|R_2).P(R_2)}{P(R_1)} \quad (7)$$

For each observation $i$ with risk factors $\{R_1, ..., R_k\}$, NB predicts the label $s_i$ in a way that the conditional probability $P(s_i|R_1, ..., R_k)$ stated in Equation (8) is the maximum.

$$P(s_i|R_1, ..., R_k) = \frac{P(R_1, ..., R_k|s_i).P(s_i)}{P(R_1, ..., R_k)} \quad (8)$$

NB makes the following assumption on the Bayes' theorem stated in Equation (8): 1) each risk factor in the dataset is conditionally not dependent on any other risk factor for a given stroke outcome (Equation 9) and 2) neglecting the denominator in Equation (7) as it will have no impact on the conditional probability. Consequently, the conditional probability of $s_i$ can be computed using Equation (10).

$$P(R_1, ..., R_k|s_i) = \prod_{j=1}^{k} P(R_j|s_i) \quad (9)$$

$$P(s_i|R_1, ..., R_k) = P(s_i). \prod_{j=1}^{k} P(R_j|s_i) \quad (10)$$

*E. Logistic Regression*

LR predicts the probability of class label $s_i$ given the risk factors $R_i$ for an observation $i$ as stated in Equation (11).

$$P(s_i) = \frac{e^{(\beta_0 + \beta_1 R_i)}}{1 + e^{(\beta_0 + \beta_1 R_i)}} \quad (11)$$

where $\beta_0$ and $\beta_1$ are the regression coefficients whose values are obtained in a way that the log loss function stated in Equation (12) is optimized. To reduce variance and overfitting in LR, a penalty term is added to the loss function as stated in Equation (13). LR consists of two hyperparameters: 1) regularization parameter and 2) solver used in the optimization problem.

$$L_{log} = -\sum_{i=1}^{n} -\ln(1 + e^{(\beta_0+\beta_1 R_i)}) + s_i^{actual}(\beta_0 + \beta_1 R_i) \quad (12)$$

$$L_{log} + \lambda \sum_{i=1}^{n} \beta_i^2 \quad (13)$$

## V. PERFORMANCE ANALYSIS

### A. Experimental Environment

To evaluate the performance of the most used five machine learning algorithms, i.e., DT, RF, SVM, NB, and LR, within the proposed framework, an imbalanced stroke prediction dataset[4] is used. Figure 3 shows the correlation between the features and the prevalence of stroke. As shown in the figure, age has the highest positive correlation with the prevalence of stroke followed by heart disease. Average glucose level, hypertension, marital status, BMI, and gender show a slight positive correlation with stroke. The residence type has almost no correlation with stroke, whereas work type and smoking status are slightly negatively correlated. As shown in Figure 4, BMI and smoking status have 1462 (3.37%) and 13292 (30.63%) observations with missing values respectively.

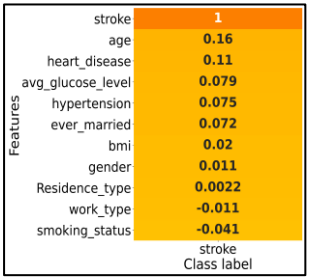 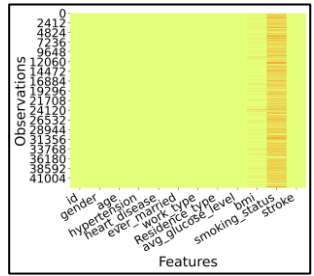

Fig 3. Correlation between the dataset features and prevalence of stroke.

Fig. 4. Distribution of missing values across observations in the dataset.

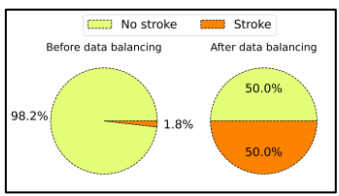

Fig. 5. Percentage of stroke and no stroke observations in the dataset before and after data balancing.

### B. Experiments

*1) Data preprocessing:* We preprocess the dataset by removing the irrelevant feature, i.e., 'ID'. To handle missing values, the observations having null values for BMI and smoking status features are removed. The categorical features, i.e., gender, ever married, work type, residence type, and smoking status, are converted to numeric using label encoding. This is by assigning a numeric value between 0 and $v-1$ to each distinct value of a feature, where $v$ represents the number of unique values for the feature. For instance, after performing label encoding on the 'gender' feature, the categorical values female, male, and other, are converted to 0, 1, and 2 respectively. The preprocessed dataset contains 10 features, 1 class label (stroke/no stroke), and 29072 observations, and is highly imbalanced as shown in Figure 5, i.e., 548 and 28,524 observations belong to stroke and no stroke classes respectively. Consequently, the minority (stroke) class is oversampled using Synthetic Minority Oversampling Technique (SMOTE) [12]. The selection of SMOTE is based on its performance compared to other data balancing approaches [16]. The dataset after balancing has 57048 observations (28524 strokes and 28524 no-stroke) as shown in Figure 5.

*2) Hyperparameter tuning:* To obtain the optimal parameters for each algorithm under study, we perform a grid search over the different values of parameters. Table II presents the hyperparameters for each algorithm, the corresponding ranges used in our experiments and literature, and the total number of combinations for the grid search. For parameters with numerical values, we select the range in a way that includes the ranges considered in the literature. For each algorithm, the parameters resulting in the maximum value of macro-averaged F-measure are considered for model building. F-measure is used for scoring as it can reveal how much a classification model is capable to identify the minority (stroke) class for an imbalanced dataset [12], [14]. All the experiments are performed using Python 3.8[7]. Default python library values are considered for the parameters not listed in the table.

*3) Model building and evaluation:* To develop and validate the model, 10-fold cross-validation is used. We evaluate the performance of each model before and after data balancing in terms of accuracy (Equation 14), precision (Equations 15-17), recall (Equations 18-20), F-measure (Equations 21-23), Receiver Operating Characteristic (ROC) curve, Area under the ROC Curve (AUC), and execution times for hyperparameter tuning and model development and validation.

$$Accuracy = \frac{TP + TN}{TP + FP + TN + FN} \quad (14)$$

$$P_{stroke} = \frac{TP}{TP + FP} \quad (15)$$

$$P_{no\ stroke} = \frac{TN}{TN + FN} \quad (16)$$

$$P_{MA} = \frac{P_{stroke} + P_{no\ stroke}}{2} \quad (17)$$

$$R_{stroke} = \frac{TP}{TP + FN} \quad (18)$$

$$R_{no\ stroke} = \frac{TN}{TN + FP} \quad (19)$$

$$R_{MA} = \frac{R_{stroke} + R_{no\ stroke}}{2} \quad (20)$$

$$F_{stroke} = \frac{2(P_{stroke} \times R_{stroke})}{P_{stroke} + R_{stroke}} \quad (21)$$

$$F_{no\ stroke} = \frac{2(P_{no\ stroke} \times R_{no\ stroke})}{P_{no\ stroke} + R_{no\ stroke}} \quad (22)$$

---

[7] https://devdocs.io/python~3.8/ (last accessed on April 25, 2022)

TABLE II. VALUE(S) OF HYPERPARAMETERS USED IN THE LITERATURE AND OUR EXPERIMENTS FOR THE ALGORITHMS UNDER STUDY.

| Algorithm | Hyperparameters | Value(s) used in the literature | Value(s) used in our experiments | Total combinations |
|---|---|---|---|---|
| Decision Tree | Splitting criteria | entropy [3], [4] and NR [5], [6], [8] | Gini and entropy | 8 |
| | Maximum features | NR [3]–[6], [8] | None, auto, sqrt, and log2 | |
| Random Forest | Number of estimators/trees | 100 [4], 500 [5], and NR [6], [7] | 10, 20, 30, 40, 50, 60, 70, 80, 90, 100, 200, 300, 400, and 500 | 112 |
| | Splitting criteria | NR [4]–[7] | Gini and entropy | |
| | Maximum features | | None, auto, sqrt, and log2 | |
| Support Vector Machine | kernel | Linear [3], Radial Basis Function [4], [5], and NR [7] | Radial Basis Function† | 30 |
| | Regularization parameter | 0.1 [4], {$2^{-2}, 2^0, 2^2, 2^4$, and $2^6$} [5], and NR [3], [7] | 0.1, $2^{-2}, 2^0, 2^2, 2^4$, and $2^6$ | |
| | Kernel coefficient | {$2^{-6}, 2^{-4}, 2^{-2}, 2^0$, and $2^2$} [5] and NR [3], [4], [7] | $2^{-6}, 2^{-4}, 2^{-2}, 2^0$, and $2^2$ | |
| Logistic Regression | Regularization parameter | {$2^{-6}, 2^{-4}, 2^{-2}, 2^0, 2^2, 2^4$, and $2^6$} [5] and NR [4], [6] | $2^{-6}, 2^{-4}, 2^{-2}, 2^0, 2^2, 2^4$, and $2^6$ | 35 |
| | Solver | NR [4]–[6] | Newton-cg, lbfgs, liblinear, sag, and saga | |
| | Maximum iterations | | 3000‡ | |

†Linear kernel is not considered as the relationship between the features and stroke is not linear; ‡The algorithm did not converge before 3000 iterations; NR: Not Reported

$$F_{MA} = \frac{2(P_{MA} \times R_{MA})}{P_{MA} + R_{MA}} \qquad (23)$$

where TP (True Positive) and TN (True Negative) indicate the number of observations correctly classified as positive (stroke) class and negative (no stroke) class respectively. FP (False Positive) and FN (False Negative) indicate the number of observations incorrectly classified as positive class and negative class respectively. $P_{stroke}$, $P_{no\ stroke}$, and $P_{MA}$ represent stroke class, no stroke class, and macro average precisions respectively. Similarly $R_{stroke}$, $R_{no\ stroke}$, and $R_{MA}$ represent the stroke class, no stroke class, and macro average recalls respectively, and $F_{stroke}$, $F_{no\ stroke}$, and $F_{MA}$ indicates the stroke class, no stroke class, and macro average F-measures respectively.

*C. Experimental Results Analysis*

Table III shows the optimal values of hyperparameters obtained for the algorithms before and after data balancing. Figure 6 depicts the accuracy obtained by the algorithms for stroke prediction. It shows that RF and LR have the highest accuracy of 98%, while NB has the least accuracy of 92% compared to other algorithms. The accuracy of DT, RF, and SVM does not change after data balancing. On the other hand, for NB and LR, the accuracy decreased after data balancing. However, the confusion matrix reveals that DT, RF, SVM, and LR are not capable of predicting the minority (stroke) class before data balancing. In particular, DT, RF, and SVM are able to correctly predict 5.5% (Figure 7(a)), 0.3% (Figure 7(b)), and 4.3% (Figure 7(c)) of the minority class respectively. In addition, LR classifies all the observations as the majority (no stroke) class (Figure 7(e)). In contrast, NB can predict 26.2% of the minority (stroke) class correctly as shown in Figure 7(d). Consequently, DT, RF, SVM, and LR result in high accuracy before data balancing compared to NB (Figure 6) as they classify almost all observations into (majority) no stroke class, leading to a high TN value without being able to predict the minority class. However, after data balancing, the considered algorithms can detect both stroke and no stroke classes as shown in Figures 7(f) – 7(j). In particular, DT, RF, and SVM can correctly predict 97.2% (Figure 7(f)), 98.4% (Figure 7(g)), and 96.6% (Figure (h)) of the stroke class respectively. On the other hand, NB and LR can correctly classify 80.6% (Figure (i)) and 80% (Figure (j)) of the stroke class respectively.

Figure 8(a) and Figure 8(b) show class-specific and macro average precision and recall respectively. For the algorithms under study, precision and recall values are more than 0.90 for the majority (no stroke) class. However, for the minority class, these values are less than 0.1 for DT, RF, SVM, and LR. This is because these algorithms are not able to detect more than 6% of the minority class. However, the minority class recall for NB is high as it classifies more minority (no stroke) observations correctly compare to other algorithms. The precision and recall for the minority class increase after data balancing for all the algorithms. NB and LR have the least precision and recall after data balancing. Figure 8(c) reveals that DT, RF, and SVM are the most suitable algorithm for predicting stroke after data balancing as it has a high class-specific F-measure compared to NB and LR.

TABLE III. OPTIMAL VALUES FOR HYPERPARAMETERS OBTAINED IN OUR EXPERIMENTS BEFORE AND AFTER DATA BALANCING.

| Algorithm | Hyperparameters | Optimal value | |
|---|---|---|---|
| | | Before data balancing | After data balancing |
| Decision Tree | Splitting criteria | Gini | Entropy |
| | Maximum features | sqrt | none |
| Random forest | Number of estimators/trees | 10 | 100 |
| | Splitting criteria | Gini | Entropy |
| | Maximum features | none | none |
| Support Vector Machine | Regularization parameter | $2^6$ | $2^2$ |
| | Kernel coefficient | $2^{-4}$ | $2^0$ |
| Logistic Regression | Regularization parameter | $2^6$ | $2^2$ |
| | Solver | newton-cg | lbfgs |

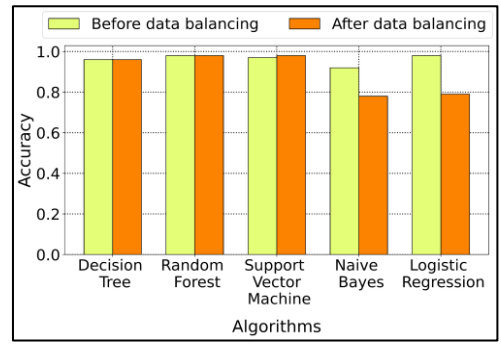

Fig. 6. Accuracy of the classification algorithms for stroke prediction before and after data balancing.

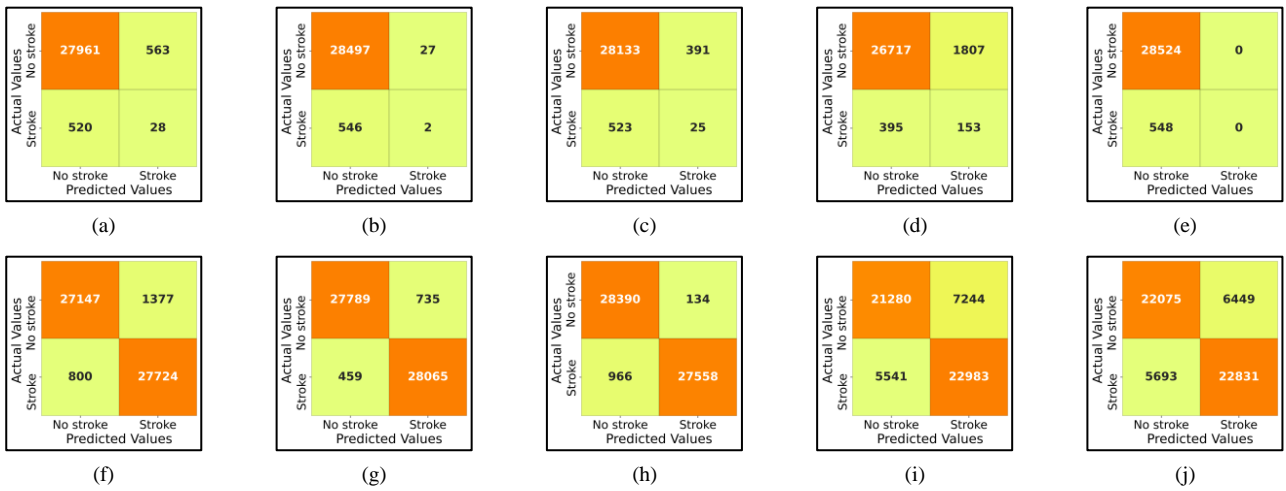

Fig. 7. Confusion Matrix: (a) Decision Tree – before data balancing, (b) Random Forest – before data balancing, (c) Support Vector Machine – before data balancing, (d) Naïve Bayes – before data balancing, (e) Logistic Regression – before data balancing, (f) Decision Tree – after data balancing, (g) Random Forest – after data balancing, (h) Support Vector Machine – after data balancing, (i) Naïve Bayes – after data balancing, and (j) Logistic Regression – after data balancing.

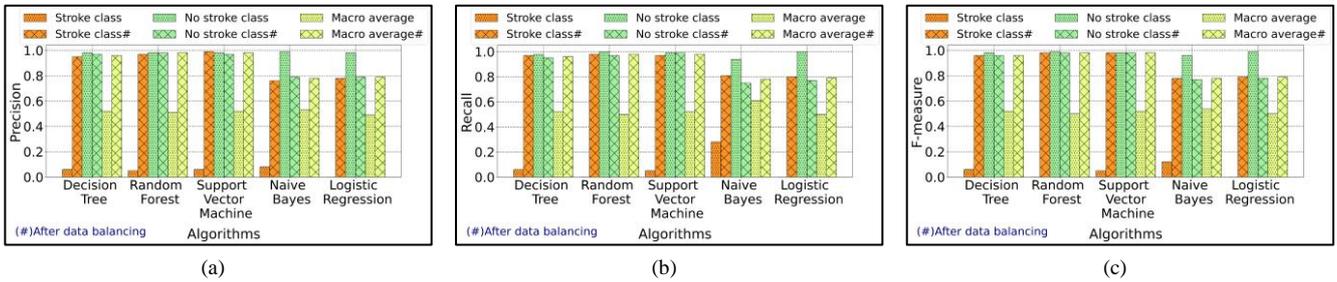

Fig. 8. Class-specific and macro average performance of the classification algorithms for stroke prediction before and after data balancing (a) precision, (b) recall, and (c) F-measure.

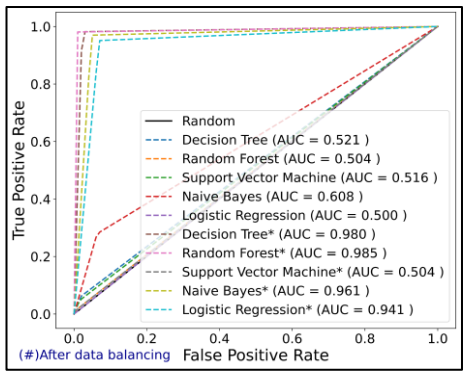

Fig. 9. Receiver Operating Characteristics (ROC) curve and Area Under the ROC Curve (AUC) of the classification algorithms for stroke prediction before and after data balancing.

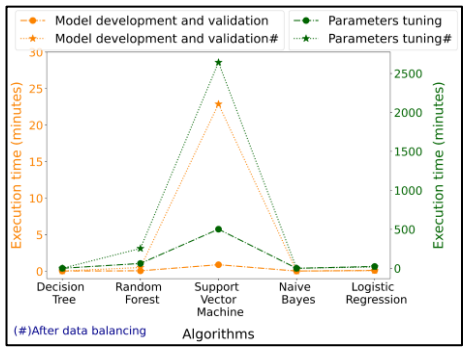

Fig. 10. Execution time of the classification algorithms for hyperparameter tuning and model development and validation before and after data balancing.

Figure 9 presents the ROC curve along with the AUC values for the algorithms. It shows that before data balancing, all the algorithms except NB has an AUC value of around 0.500, which is the same as random prediction, as these algorithms are not detecting the minority class. On the other, hand NB has the highest AUC value of 0.608 indicating the detection of both minority (stroke) and majority (no stroke) classes before data balancing. After data balancing, RF has the highest AUC value of 0.985. Figure 10 presents the execution time for parameters tuning as well as model development and validation before and after data balancing. It shows that SVM takes the highest time for parameter tuning and model development for both before and after data balancing, whereas NB takes the least time. The execution time increases after data balancing because of increased number of observations in the dataset due to oversampling of the minority class.

In summary, NB has the best performance in terms of predicting minority class before data balancing and RF has the best prediction performance after data balancing.

## VI. CONCLUSIONS

Being the second leading cause of death and a global crisis, it is crucial to predict the prevalence of stroke in an individual based on risk factors. An intelligent stroke prediction framework is proposed to provide prognosis and diagnosis support to allied health professionals and to aid health stakeholders to develop a nationwide stroke prevention plan. In addition, the most five used machine learning algorithms for stroke prediction are evaluated and compared, using a unified setup, in terms of accuracy, class-specific and macro average precision, recall and F-measure, ROC, AUC, and execution times for hyperparameters tuning and model development and validation. In the future, a large spectrum of machine learning models will be considered. In addition, the performance of deep learning models for stroke prediction will be studied.

ACKNOWLEDGMENT

This research was funded by the National Water and Energy Center of the United Arab Emirates University (Grant 31R215).